%% file: main.tex
\documentclass[10pt,twocolumn,letterpaper]{article}

\usepackage{iccv}
\usepackage{times}
\usepackage{epsfig}
\usepackage{graphicx}
\usepackage{amsmath}
\usepackage{amssymb}

\usepackage{multirow}
\usepackage{makecell}
\usepackage[table]{xcolor}
\usepackage{amsfonts}
\usepackage{bbm}
\usepackage{commath}
\usepackage{paralist}
\usepackage{mathtools}
\usepackage{nccmath}
\usepackage{pifont}
\usepackage{tabularx}
\usepackage{makecell}
\usepackage{subcaption}
\usepackage[accsupp]{axessibility}

\newcolumntype{g}{>{\columncolor{green!50}}c}
\newcolumntype{y}{>{\columncolor{yellow!50}}c}
\newcolumntype{o}{>{\columncolor{orange!50}}c}
\newcolumntype{e}{>{\columncolor{red!50}}c}

\newcommand{\cmark}{\ding{51}}%
\newcommand{\xmark}{\ding{55}}%

\usepackage[pagebackref=true,breaklinks=true,colorlinks,bookmarks=false]{hyperref}

\usepackage[capitalize]{cleveref}
\crefname{section}{Sec.}{Secs.}
\Crefname{section}{Section}{Sections}
\Crefname{table}{Table}{Tables}
\crefname{table}{Tab.}{Tabs.}

\iccvfinalcopy 

\def\iccvPaperID{****} 

\begin{document}

\title{Deformer: Dynamic Fusion Transformer for Robust Hand Pose Estimation}

\author{Qichen Fu\textsuperscript{\rm 1}
\qquad
Xingyu Liu\textsuperscript{\rm 1}
\qquad
Ran Xu\textsuperscript{\rm 2}
\qquad
Juan Carlos Niebles\textsuperscript{\rm 2}
\qquad
Kris M. Kitani\textsuperscript{\rm 1} \\
\textsuperscript{\rm 1} Carnegie Mellon University
\qquad
\textsuperscript{\rm 2} Salesforce Research
\and
\small \url{https://fuqichen1998.github.io/Deformer/}
}

\maketitle

\begin{abstract}
\input{abstract}
\end{abstract}

\section{Introduction}
\label{sec:intro}
\input{introduction}

\section{Related Work}
\label{sec:related-work}
\input{related-work}

\section{Method}
\label{sec:method}
\input{method}

\section{Experiments}
\label{sec:exp}
\input{experiments}

\section{Conclusion}
\label{sec:conclusion}
\input{conclusion}

\noindent\textbf{Acknowledgement: } This work is funded in part by JST AIP Acceleration (Grant Number JPMJCR20U1, Japan), and Salesforce.

\appendix
\input{supplementary.tex}

\clearpage
{\small
\bibliographystyle{ieee_fullname}
\bibliography{egbib}
}

\end{document}

%% file: abstract.tex
Accurately estimating 3D hand pose is crucial for understanding how humans interact with the world. Despite remarkable progress, existing methods often struggle to generate plausible hand poses when the hand is heavily occluded or blurred. In videos, the movements of the hand allow us to observe various parts of the hand that may be occluded or blurred in a single frame. To adaptively leverage the visual clue before and after the occlusion or blurring for robust hand pose estimation, we propose the Deformer: a framework that implicitly reasons about the relationship between hand parts within the same image (spatial dimension) and different timesteps (temporal dimension). We show that a naive application of the transformer self-attention mechanism is not sufficient because motion blur or occlusions in certain frames can lead to heavily distorted hand features and generate imprecise keys and queries. To address this challenge, we incorporate a Dynamic Fusion Module into Deformer, which predicts the deformation of the hand and warps the hand mesh predictions from nearby frames to explicitly support the current frame estimation. Furthermore, we have observed that errors are unevenly distributed across different hand parts, with vertices around fingertips having disproportionately higher errors than those around the palm. We mitigate this issue by introducing a new loss function called maxMSE that automatically adjusts the weight of every vertex to focus the model on critical hand parts. Extensive experiments show that our method significantly outperforms state-of-the-art methods by 10\%, and is more robust to occlusions (over 14\%).

%% file: introduction.tex
\begin{figure}[t]
	\centering
	\includegraphics[width=\linewidth]{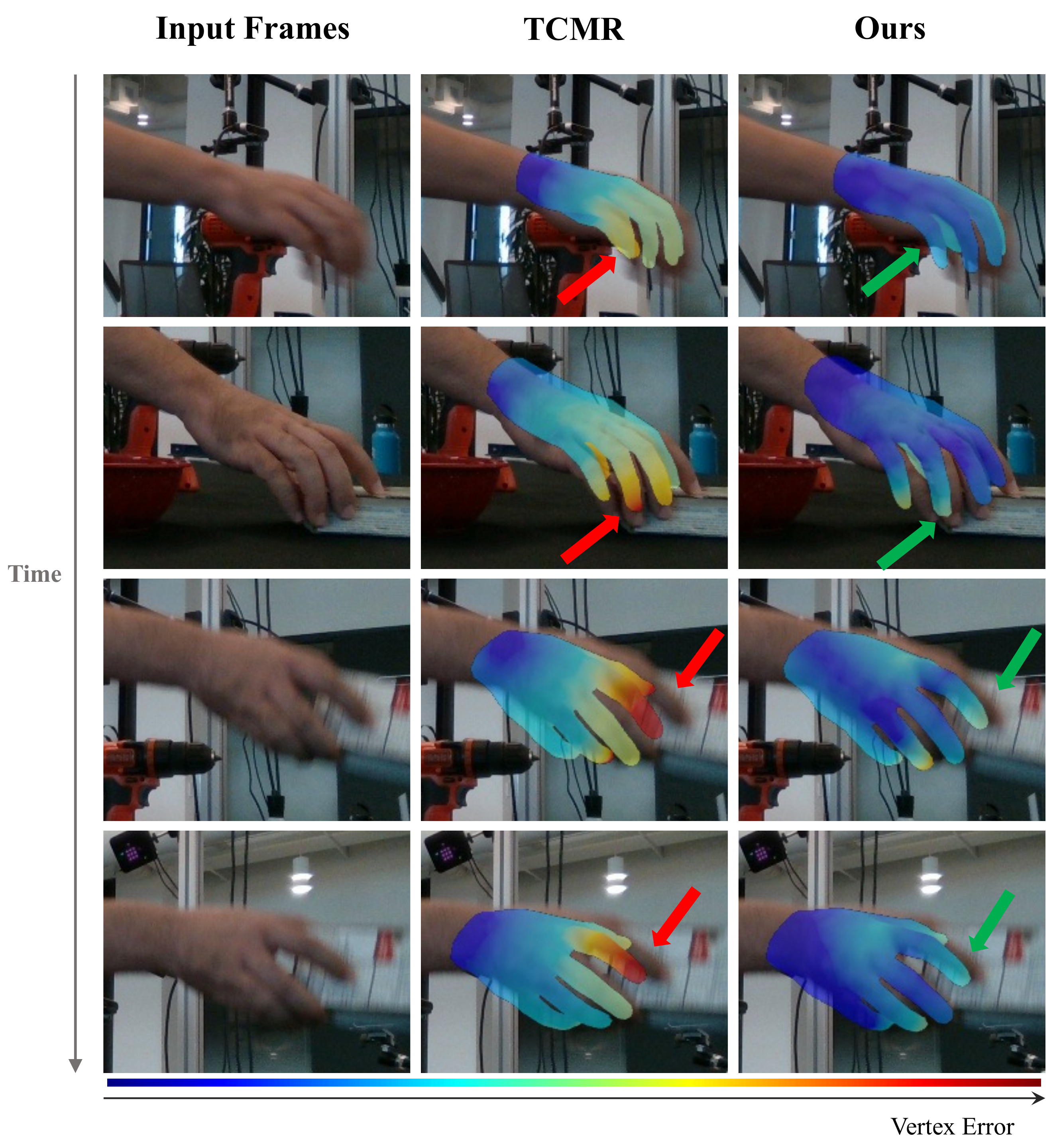}
	\caption{Given a video where in some frames (\textit{left}) the hand is heavily occluded or blurred, the existing state-of-the-art video-based method TCMR \cite{tcmr} (\textit{middle}) fails to predict accurate hand poses. Our method (\textit{right}) is able to capture the hand dynamics and leverage neighborhood frames to robustly produce plausible hand pose estimations.}
	\label{fig:teaser}
\end{figure}

Accurately estimating hand poses in the wild is a challenging task that is affected by various factors, including object occlusion, self-occlusion,  motion blur, and low camera exposure. 
To address these challenges, temporal information can be leveraged using a self-attention mechanism \cite{transformer} that reasons the feature correlation between adjacent frames to generate better hand pose estimations. However, as the generation of key and query vectors depends on per-frame features, heavy occlusion or blurring can significantly contaminate them, resulting in inaccurate output. Another challenge of hand pose estimation is that the fingertips, located at the periphery of the hand, are more prone to occlusion and have more complex motion patterns, making them particularly challenging for the model to estimate accurately.

To tackle the aforementioned challenges, we propose Deformer: a dynamic fusion transformer that leverages nearby frames to learn hand deformations and assemble multiple wrapped hand poses from nearby frames for a robust hand pose estimation in the current frame. Deformer implicitly models the temporal correlations between adjacent frames and automatically selects frames to focus on. To mitigate the error imbalance issue, we design a novel loss function, called maxMSE, that emphasizes the importance of difficult-to-estimate vertices and provides a more balanced optimization.

Given a sequence of frames, our approach first uses a shared CNN to extract frame-wise hand features. To reason the non-local relationships between hand parts within a single frame, we use a shared spatial transformer that outputs an enhanced per-frame hand representation. Then, we leverage a global temporal transformer to attend to frame-wise hand features by exploring their correlations between different timestamps and forward the enhanced features to a shared MLP to regress the MANO hand pose parameters. To ensure consistency of the hand shape over time, we predict a global hand shape representation from all frame-wise features through the cross-attention mechanism. Despite incorporating temporal attention, the model may struggle to accurately predict hand poses in frames where the hand is heavily occluded or blurred. To address this issue, we introduce a \textit{Dynamic Fusion Module}, which predicts a tuple of forward and backward hand motion for each frame, deforming the hand pose to the previous and next timestamps. This allows the model to leverage nearby frames with clear hand visibility to assist in estimating hand poses in occluded or blurred frames. Finally, the set of deformed hand poses of each timestamp is synthesized into the final output with implicitly learned confidence scores.
In optimization, we found that standard MSE loss leads to imbalanced errors between different hand parts. Specifically, we observed that the model trained with MSE loss had lower errors for vertices around the palm and larger errors for the vertices around the fingertips, which are more prone to occlusions and have more complex motion patterns. Inspired by focal loss~\cite{lin2017focal} in the classification task, we introduce \textit{maxMSE}, a new loss function that maximizes the MSE loss by automatically adding a weight coefficient to every hand vertex. The \textit{maxMSE} allows the model to focus on critical hand parts like fingertips and achieve better overall performance.

Experimental results on two large-scale hand pose estimation video datasets, DexYCB~\cite{dexycb} and HO3D~\cite{ho3d}, show the proposed method achieves state-of-the-art performance and is more robust to occlusions. In summary, our contributions include: (1) a Deformer architecture for robust and plausible 3D hand pose estimation from videos; (2) a novel dynamic fusion module that explicitly deforms nearby frames with clear hand visibility for robust hand pose estimation in occluded or blurred frames; (3) a new \textit{maxMSE} loss function that focuses the model on critical hand parts.

%% file: related-work.tex
\begin{figure*}[t]
\centering
\includegraphics[width=\linewidth]{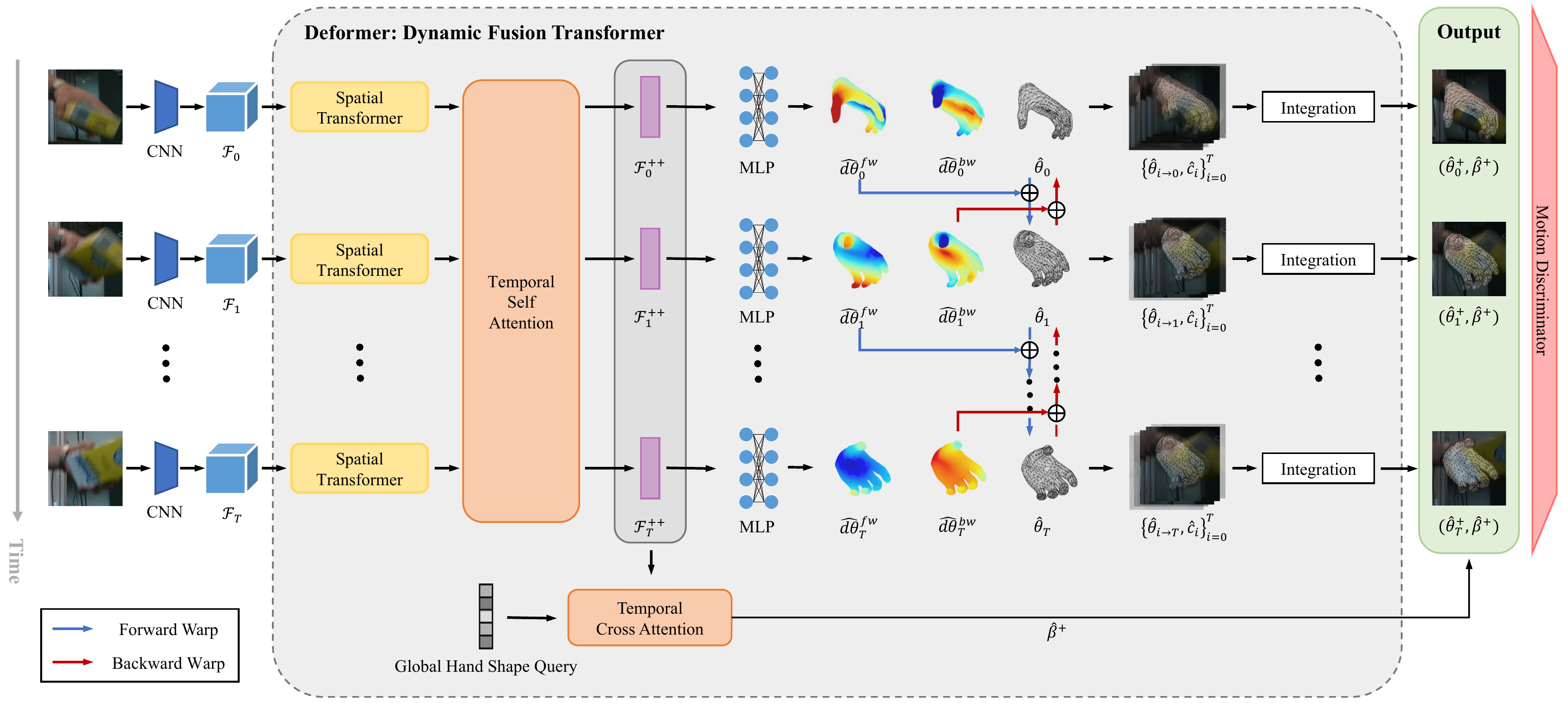}
\caption{Overview of the Deformer Architecture. Our approach uses transformers to reason spatial and temporal relationships between hand parts in an image sequence, and output frame-wise hand pose and motion. In order to overcome the challenge when the hand is heavily occluded or blurred in some frames, the Dynamic Fusion Module explicitly deforms the hand poses from neighborhood frames and fuses them toward a robust hand pose estimation.}
\label{fig:overview}
\end{figure*} 

3D hand pose estimation (HPE) aims to predict the 3D coordinates of the 21 hand joints, or more precisely, recover the hand mesh. It is a crucial computer vision task for understanding human activities, which plays an important role in many applications, including human-computer interaction~\cite{pavlovic1997visual, rautaray2015vision, von2001bare}, AR/VR~\cite{piumsomboon2013user, holl2018efficient}, imitation learning~\cite{vogt2007prefrontal, handa2020dexpilot, garcia2020physics}, and etc. 

\textbf{Hand Pose Estimation from a Single Image }
Single-view methods for hand pose estimation can be broadly divided into two categories: model-free and model-based methods.

Model-free methods~\cite{zimmermann2017learning, yang2019disentangling, lin2021end, li20143d, park20163d, pavlakos2017coarse, kolotouros2019convolutional, weaklysuphpe} directly regress the 3D coordinates or occupancy of hand joints and mesh vertices. For example, \cite{kolotouros2019convolutional} uses a graph convolutional neural network to model triangle mesh topography and directly predicts the 3D coordinates of hand mesh vertices. \cite{pavlakos2017coarse} discretizes the 3D space into a voxel grid, and proposed a CNN in the stacked hourglass~\cite{newell2016stacked} style to estimate per joint occupancy. As hand mesh has a high degree of freedom, model-free methods need to introduce biological constraints~\cite{weaklysuphpe} to get reasonable predictions. Meanwhile, these methods need a lot of data to train due to the lack of priors. 

Recently, \cite{mano} introduces a pre-defined hand model, named MANO, which contains abundant structure priors of human hands. It reduces the dimensions by mapping a set of low-dimensional pose and shape parameters to a hand mesh. Most model-based methods \cite{jointho, semihandobj, baek2019pushing, boukhayma20193d, zhang2019end} instead predict the MANO parameters, which consists of the pose and shape deformation w.r.t. a mean pose and shape learned from various hand scans of multiple subjects. 3D joints and mesh vertices coordinates can be retrieved from the predicted parameters using the differentiable MANO model. For instance, \cite{zhang2019end} develops an end-to-end framework optimized by a differentiable re-projection loss to recover the hand mesh. \cite{jointho} exploits the contact loss between hand and object to ensure the prediction is physically plausible. In this work, we leverage the MANO model to reduce reliance on annotations and prevent overfitting.

While image-based methods have made remarkable progress, their predictions are not stable due to the occlusions and visual jitters in static images.
 
\textbf{Temporal Hand Pose Estimation }
Some recent methods \cite{semihandobj, yang2020seqhand, vibe, hasson2020leveraging, handocc, fan2023arctic, zhou2022toch, tiwari2022pose} start to explore the temporal information in videos to regularize the per-frame prediction. \cite{hasson2020leveraging} exploits the photometric consistency between the re-projected 3D hand prediction and the optical flow of nearby frames. \cite{yang2020seqhand, semihandobj} impose the temporal consistency for smooth motions. \cite{vibe} introduces a recurrent network as a discriminator to supervise motion sequences through adversarial training. Though the above methods use temporal information as extra supervision, they are not specialized for sequence input and the models fail to leverage the dynamic from the video.

Alternatively, \cite{cai2019exploiting} exploits spatial and temporal relationships using a graph convolutional neural network to predict 3D hand pose. However, it relies on the 2D pose estimator \cite{newell2016stacked} to pre-computed consecutive 2D hand poses as input and can't be optimized jointly. Recently, \cite{multiviewvideohpe} utilizes temporal information by propagating frame-wise features using a sequence-to-sequence model based on LSTM \cite{lstm}, yet it only gives sparse joint location predictions. \cite{tcmr} presents a temporally consistent system that synthesizes the implicit feature from the past and future to recover the mesh of the center frame. Those methods, mainly relying on recurrent neural network, assumes a strong dependency on local relationships, which suffers from modeling long-range dependency. Differently, our transformer-based method can explicitly attend to every frame as direct evidence, without being constrained by spatial and temporal gaps, to give an accurate hand pose estimation.

\textbf{Transformer-based Methods }
Transformer \cite{transformer} has been the dominant model in various NLP tasks \cite{devlin2018bert, radford2018improving, radford2019language}. In contrast to CNNs, the self-attention mechanism at the heart of the Transformer has no strong inductive biases and can adaptively attend to a sequence of features. Inspired by its success and versatility, there are growing interests in applying the self-attention layer to computer vision tasks, including classification \cite{vit, liu2021swin}, detection \cite{detr, zhu2020deformable}, video understanding \cite{arnab2021vivit, liu2022video}, human-object interaction \cite{fu2022sequential, kim2021hotr}, etc. Recently, \cite{lin2021end, huang2020hot} applied the transformer to exploit non-local interactions for 3D human pose and mesh reconstruction from a single image. \cite{semihandobj} utilizes the transformer to perform explicit contextual reasoning between hand and object representations. Differently from these methods only focusing on the spatial relationships in static images, we propose a sequential transformer structure to simultaneously model spatiotemporal relationships of hand joints and their motion for hand pose estimation from video.

%% file: method.tex
The overview structure of our framework is illustrated in \cref{fig:overview}. Given an input image sequence $V=\{I_t\}_{t=1}^T$ of length T, we aim to estimate the 3D hand joints $\mathcal{J}_t$ and mesh vertices $\mathcal{V}_t$. To achieve the goal, we first use a pretrained CNN to extract the hand feature map in every frame. Then we learn transformers to sequentially reason spatial and temporal relationships, which outputs a set of attended latent vectors containing information on hand mesh and motion at each timestamp. The latent feature vector, on one hand, is used to regress the MANO parameters by an MLP, which is served as the input to the differential MANO\cite{mano} hand layer to recover the hand mesh.

In addition to estimating MANO parameters, we also proposed a dynamic fusion module to estimate deformation parameters and a confidence score from the hidden feature to model hand motion and frame accountability. Finally, the predicted hand motions are utilized to explicitly deform the hand mesh prediction of all timestamps into other frames and integrated by the implicitly learned confidence scores, resulting in a more accurate estimation of hand pose and better handling of occlusions and blurs.

Apart from the MANO parameters, we further estimate the deformation parameters and a confidence score from the hidden feature to model the motion of the hand and the accountability of the frame. Furthermore, we utilize the predicted hand motions to explicitly deform the hand mesh prediction of all timestamps into other frames and integrate them with the implicitly learned confidence scores. This results in a more precise estimation of hand pose and better handling of occlusions and blurs.

\subsection{Deformer}

The Deformer is a sequential combination of a spatial transformer, a temporal transformer, and a dynamic fusion module. The spatial transformer focuses on extracting a compact representation from an image, which is robust to occlusions. The network weight of the spatial transformer is shared for all frames. The temporal transformer enhances the hand feature by exploiting the correlation of hand in every timestamp and regresses a global hand shape. The dynamic module learns the hand motion and frames accountability to explicitly fuse the frame-wise predictions toward final hand pose estimations in a confidence-driven manner.

\textbf{Spatial Transformer }
Following the idea of the hybrid model \cite{detr, kim2021hotr}, our method first uses the truncated FPN~\cite{fpn} with ResNet50~\cite{resnet}, followed by ROIAlign~\cite{he2017mask} to extract initial hand feature $\mathcal{F}_t \in \mathcal{R}^{H\times W\times C}$ for every timestamp $t$. The feature extracted by CNN, fundamentally using a set of convolutional kernels, has an inductive bias toward two-dimensional neighborhood structure. It suffers from occlusions where local structures are contaminated. In order to overcome this challenge, we incorporate the Transformer~\cite{transformer} to address non-local relationships to make our method more robust.

The spatial transformer encoder expects a sequence as input. Thus, we flatten the spatial dimension of the hand feature, resulting in $HW$ vectors where each has a length of $C$. Those feature vectors, named tokens, are processed by several layers of multi-head self-attention and feed-forward network (FFN). Since every layer of the transformer is permutation-invariant, we supplement the 2D location information to every token by adding 2D positional embeddings. The output feature $\mathcal{F}_t^e\in\mathcal{R}^{HW\times C}$ of the transformer encoder reinforces the original feature by capturing the non-local interactions of all tokens. The enhanced feature should contain rich contextual about the hand, including the location of hand joints. As an intermediate supervision, we regress a heatmap of hand joints $\hat{\mathcal{H}}_t$ from $\mathcal{F}^e_t$, and retrieve the $N_j=21$ joints location prediction $\hat{\mathcal{J}}_t^{2D}\in\mathcal{R}^{21\times 2}$. Inspired by the idea of skip connection~\cite{resnet}, the joints heatmap is explicitly added to $\mathcal{F}_t^e$ through concatenation.

The spatial transformer decoder takes a learnable query vector $\mathcal{Q}\in\mathcal{R}^C$ as input, and gradually fuses the information from the combined feature of $\mathcal{F}_t^e$ and $\hat{\mathcal{J}}^{2D}$ into it through multiple layers of cross-attention and FFN. Different from self-attention generating query, key, and value from the same set of tokens, cross-attention uses the learnable query vector to generate the query and utilizes learned tokens as the source for key and value. Similar to \cite{jaegle2021perceiver}, the decoder adaptively selects relevant information from the feature map and outputs a compact latent representation $\mathcal{F}_t^+\in\mathcal{R}^C$ of 3D hand. This greatly reduces the computational cost of temporal reasoning in the following stage.

\textbf{Temporal Transformer }
Estimating hand pose simply based on a single image is not robust for at least two reasons. First, the visual appearance of the hand may change significantly as time proceeds, causing image-based models to generate unstable predictions. Second, a hand could be frequently occluded by itself or objects during hand-object interaction. In these scenarios, an accurate hand pose estimation from a static image is even more challenging as only part of the hand is visible.

In a video, movements of the hand allow us to observe various parts of the hand which may be occluded or blurred in a single frame. Motivated by this, we use a transformer encoder to reason the feature interaction along the temporal dimension. Different from \cite{multiviewvideohpe, tcmr, vibe} using recurrent architecture \cite{lstm, gru}, the self-attention mechanism of the transformer allows every frame directly benefits from other frames without being constrained by the temporal distance. Similar to the spatial transformer encoder, the temporal transformer encoder accepts the sequence of features of every frame $\{\mathcal{F}_t^+\}_{t=1}^T$ and outputs a new set of latent vectors $\{\mathcal{F}_t^{++}\}_{t=1}^T$. Given the feature, we use an MLP as the mesh regression network to estimate the pose parameter $\hat{\theta}_t$. As the hand shape is consistent throughout all time, we use a standard transformer decoder that extracts a single feature from all frame-wise features and predicts a global shape parameter $\hat{\beta}^+$.

\begin{figure}[t]
    \centering
    \includegraphics[width=0.95\linewidth]{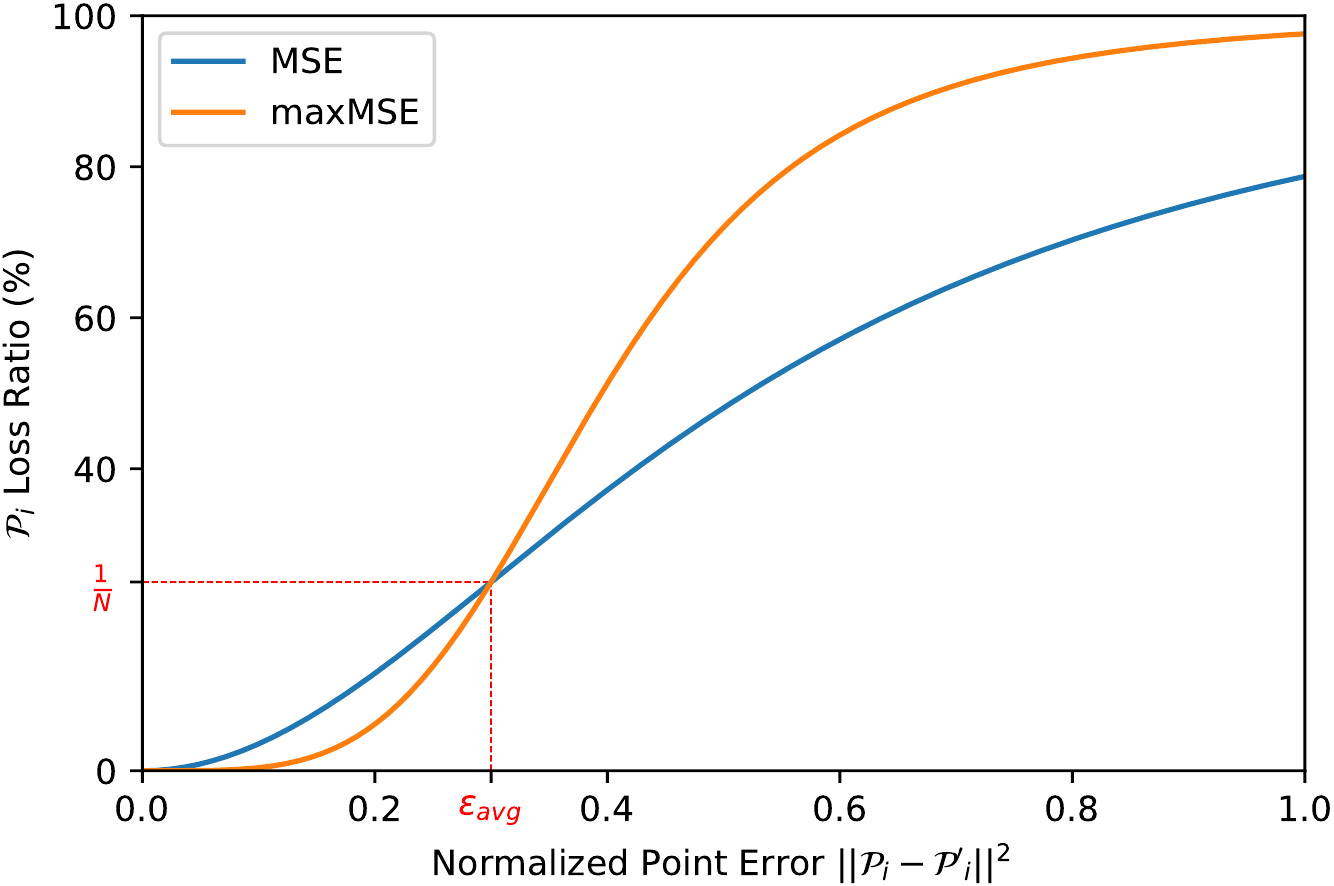} \\
    \includegraphics[width=0.59\linewidth]{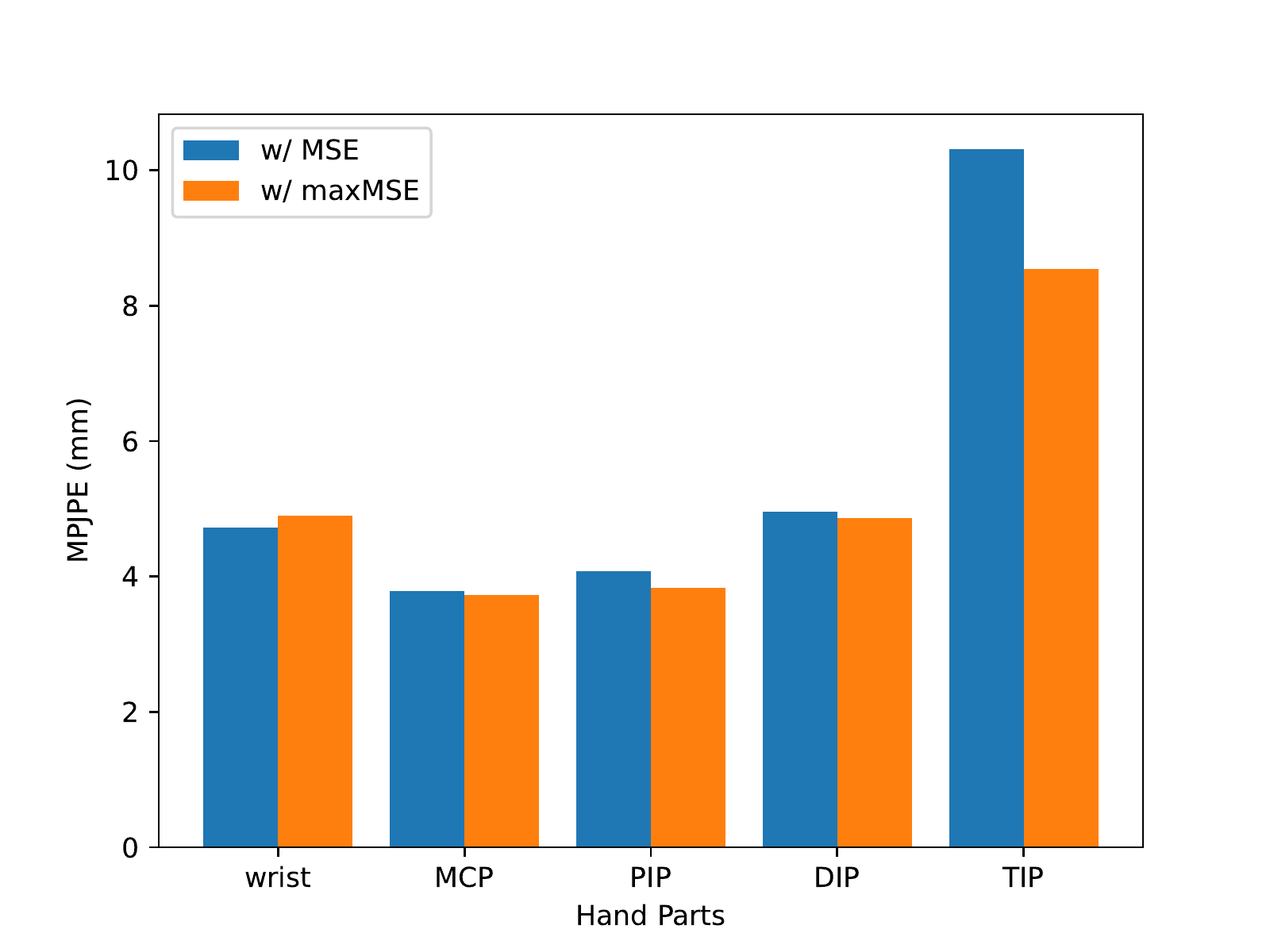}
    \includegraphics[width=0.39\linewidth]{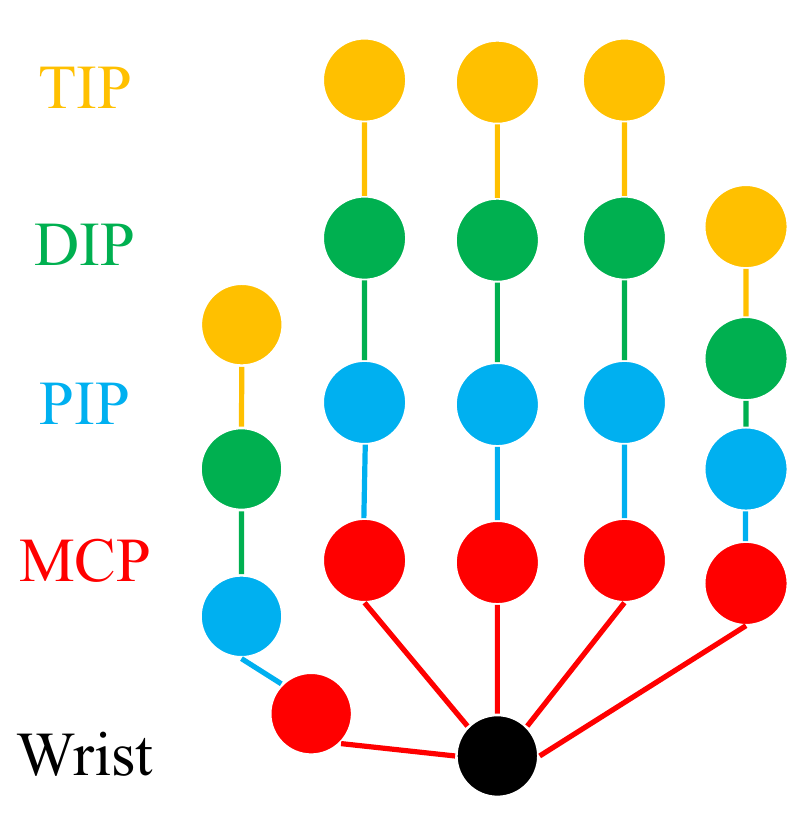}\\
    \caption{We propose a novel maxMSE loss function \textit{(top)} that automatically assigns larger weights to the hand parts with more significant errors (around fingertips), and lower weights for well-predicted parts (around the wrist). Experiments demonstrate the proposed maxMSE loss mitigates the error imbalance issue \textit{(bottom)} and leads to better overall performance.}
    \label{fig:mse_vs_maxmse}
\end{figure}

\begin{figure*}[t]
\centering
\includegraphics[width=\linewidth]{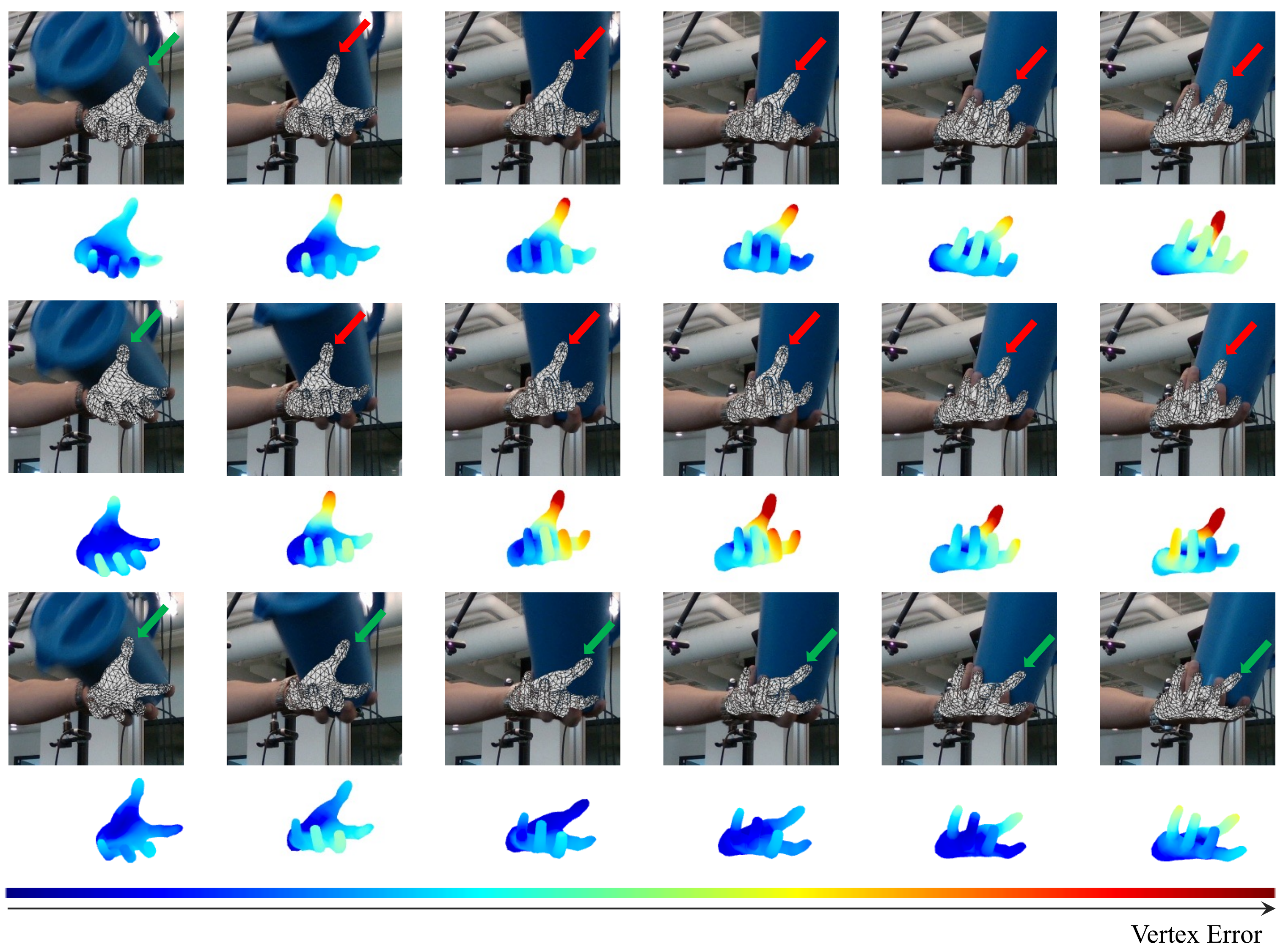}
\caption{Qualitative Comparison of the state-of-the-art single-view method~\cite{semihandobj} \textit{(top)}, video-based method~\cite{tcmr} \textit{(middle)}, and our method \textit{(bottom)}. For each method, the first row shows the predicted hand mesh, and the second row shows the error map. The proposed Deformer robustly generates more stable and accurate poses over time than previous approaches.}
\label{fig:qualitative_comparsion}
\end{figure*} 

\textbf{Dynamic Fusion Module }
Though the output feature of the temporal encoder encodes the relationships between the neighborhood frames, it still has a strong dependency on the current frame. When the current frame is heavily occluded or blurred, we found the model remains challenging to regress an accurate hand mesh. However, the predictions of its neighborhood are decent because the hand is more clearly visible in those frames. Motivated by this observation, we argue it is important to refer to other frames to help for a robust estimation of the current frame.

To achieve this goal, we predict a confidence score $\hat{c}_t$, and a tuple of forward and backward motion $(\hat{d\theta}_t^{fw}, \hat{d\theta}_t^{bw})$ as extra output from temporally attended latent vector $\mathcal{F}_t^{++}$ for every frame. The motion ground truth is the deformation of MANO parameters to the next and previous frames.

\begin{equation}
    d\theta_t^{fw} = \theta_{t+1} - \theta_t, \quad d\theta_t^{bw} = \theta_{t-1} - \theta_{t}
\end{equation}

The frame-wise motion prediction allows us to deform the predicted hand pose of frame $i$ to any frame $j$ by
\begin{equation}
    \hat{\theta}_{i\to j} =  \begin{cases}
     \hat{\theta}_i + \sum_{k=j}^{i-1} \hat{d\theta}_k^{bw} & \text{if } j < i\\
     \hat{\theta}_i + \sum_{k=i}^{j-1} \hat{d\theta}_k^{fw} & \text{if } j > i\\
    \end{cases}
\end{equation}
We deform the hand pose of all frames in the sequence to every timestamp $t$ and synthesize them. Formally, the final hand pose prediction $\hat{\theta}_t^+$ is the weighted sum of deformed frame-wise predictions based on the confidence
\begin{equation}
    \hat{\theta}_t^+ =\frac{\sum_{t'=1}^T e^{\hat{c}_{t'}} \hat{\theta}_{t'\to t} }{\sum_{t'=1}^T e^{\hat{c}_{t'}}} 
\end{equation}

The 3D coordinates estimation of hand joints $\hat{\mathcal{J}}_t\in\mathcal{R}^{21\times3}$ and mesh vertices $\hat{\mathcal{V}}_t\in\mathcal{R}^{778\times3}$ can be recovered from the predicted MANO parameters $(\hat{\theta}_t^+, \hat{\beta}^+)$ with the differentiable MANO layer~\cite{mano}.

\subsection{Optimization}

As the Deformer gives a sequence of hand poses as output, its training is ill-posed without proper regularization. To this end, we first design a novel loss function, named maxMSE, to supervise the frame-wise hand mesh output. In order to address the complex hand dynamics, we further add temporal constraints and a learnable motion discriminator. The total loss is the linear combination of them

\begin{equation}
    \mathcal{L}_{hand} = \mathcal{L}_{mesh} + \mathcal{L}_{adv} + \mathcal{L}_{aux}
\end{equation}

\textbf{maxMSE Loss }
MSE is a standard loss function for the 3D joints and mesh vertices locations. However, we found there is a large discrepancy of error between different hand positions. For example, the predictions of fingertips generally have large errors than those around the palm. This fact motivated us to design a loss that adjusts the weight for different hand parts. Inspired by focal loss~\cite{lin2017focal}, we introduce the maxMSE loss, which adjusts the weight of every location $w_i=\frac{(\mathcal{P}_i-\mathcal{P}'_i)^2}{{\sum_{j=1}^N(\mathcal{P}_j-\mathcal{P}'_j)^2}}$ to maximize the MSE loss. Specifically, the maxMSE loss for two sets of 3D points $\mathcal{P}, \mathcal{P}' \in\mathcal{R}^{N\times3}$ is

\begin{align}
    \text{maxMSE}(\mathcal{P}, \mathcal{P}') = \sum_{i=1}^N w_i ||\mathcal{P}_i-\mathcal{P}'_i||^2 = \frac{\sum_{i=1}^N||\mathcal{P}_i-\mathcal{P}'_i||^4}{\sum_{i=1}^N||\mathcal{P}_i-\mathcal{P}'_i||^2}
\end{align}

In training, maxMSE loss assigns a large weight to the joints or vertices which have larger errors. It allows the model to focus on difficult parts of the hand. We use maxMSE as the loss of predicted hand mesh as

\begin{equation}
    \mathcal{L}_{mesh} = \sum_{t=1}^T \text{maxMSE}(\hat{\mathcal{V}}_t, \mathcal{V}_t) + \text{maxMSE}(\hat{\mathcal{J}}_t, \mathcal{J}_t)
\end{equation}

Note we do not have a ground truth for the confidence score of each frame $\hat{c}_t$.  However, since it is compounded with the final hand pose prediction, the end-to-end loss $\mathcal{L}_{hand}$ implicitly supervises the confidence score.

\textbf{Motion Discrimination }
The mesh loss enforces the model to generate a hand pose aligned with the visual evidence. However, as the temporal continuity of hand dynamic is ignored, multiple inaccurate hand pose estimations in a sequence may still be recognized as valid. Following \cite{vibe}, we mitigate this issue by learning a motion discriminator $\mathcal{D}$. The motion discriminator accepts a sequence of hand mesh and tells whether it is real or not. To capture the complexity and variety of hand motion, our motion discriminator is a multi-layer GRU~\cite{gru} neural network. The model details are included in the supplementary.

We train the Deformer (generator) and the motion discriminator together using adversarial training~\cite{goodfellow2014generative}. The losses of the generator and discriminator are 
\begin{align}
    &\mathcal{L}_{adv} = \mathbb{E}_{\Theta\sim p_G}(\mathcal{D}(p_G)-1)^2 \\
    &\mathcal{L}_{\mathcal{D}} = \mathbb{E}_{\Theta\sim p_G}(\mathcal{D}(p_G))^2 + \mathbb{E}_{\Theta\sim p_R}(\mathcal{D}(p_R)-1)^2
\end{align}
where $p_G$ is a predicted hand mesh sequence and the $p_R$ is a randomly sampled ground truth sequence.

\textbf{Auxiliary Constrains }
The auxiliary loss~\cite{al2019character} has been shown to be beneficial for training a deep transformer. Our auxiliary loss comprises two components: (1) intermediate supervision to stabilize the training, and (2) temporal constraints to regularize the sequential output.

\begin{equation}
    \mathcal{L}_{aux} = \mathcal{L}_{2D} + \mathcal{L}_{monocular} + \mathcal{L}_{motion} + \mathcal{L}_{smooth}
\end{equation}

Intermediate Supervision is applied to the intermediate features $\mathcal{F}^e_t$ and $\mathcal{F}^+_t$ as each timestamp. The output feature $\mathcal{F}^e_t$ of the spatial transformer encoder should encode the 2D static hand feature. We use it to regress a heatmap of 2D hand joints and retrieve the 21 hand joints prediction $\hat{\mathcal{J}}^{2D}$. The loss of $\hat{\mathcal{J}}^{2D}$ is the difference between the hand joints prediction and the re-projected ground truth hand mesh.

\begin{equation}
    \mathcal{L}_{2D} = \sum_{j\in N_j} ||\hat{\mathcal{J}}^{2D}_j - \mathcal{J}^{2D}_j||_2^2
\end{equation}

The hand features $\mathcal{F}^e_t$ are aggregated into a single latent vector $\mathcal{F}^+_t$ by the spatial transformer decoder, which should represent the 3D hand shape and pose. To supervise this feature, we use it to estimate the MANO parameters ($\hat{\theta}_t, \hat{\beta}_t$) at every timestamp as the monocular prediction of our method. Similar to the final output, we retrieve the 3D mesh using the differential MANO layer. The monocular loss $\mathcal{L}_{monocular}$ is computed by comparing the predicted 3D mesh with the ground truth using the proposed maxMSE loss function. Note that in addition to the 3D joints and vertices, the maxMSE loss also applies to all predicted MANO parameters.

Temporal Constraints are employed to regularize the sequential output to be consistent and smooth. First, we add a motion loss $\mathcal{L}_{motion}$ to enforce the deformed hand mesh ($\hat{\theta}_{t\to t-1}$, $\hat{\theta}_{t\to t+1}$) based on the predicted motion ($\hat{d\theta}_t^{bw}$, $\hat{d\theta}_t^{fw}$) to be consistent with the previous and subsequent frames. Second, in order to generate a smooth hand pose sequence without jittering, the hand motion should be slow. We achieve this goal by adding the smooth loss $\mathcal{L}_{smooth}$ that encourages the first-order and second-order derivative of the hand pose estimations to be close to zero.

\begin{table*}[t]
\centering
\small{
\begin{tabular}{cc|cyoe}
\Xhline{1.0pt}  
Methods & Input & All & Occlusion (25\%-50\%) & Occlusion (50\%-75\%) & Occlusion (75\%-100\%)\\
\Xhline{1.0pt}
A2J~\cite{a2j} & Depth & $12.07\ (76.0)$ & $12.44\ (75.3)$ & $14.74\ (70.7)$ & $19.59\ (61.5)$ \\
\cite{weaklysuphpe} + ResNet50 & Monocular & $7.12\ (85.8)$ & $7.65\ (84.7)$ & $8.73\ (82.6)$ & $11.90\ (76.3)$ \\
\cite{weaklysuphpe} + HRNet32 & Monocular & $6.83\ (86.4)$ & $7.22\ (85.6)$ & $8.00\ (84.0)$ & $10.65\ (78.8)$ \\
MeshGraphormer~\cite{meshformer} & Monocular & $6.41\ (87.2)$ & $6.85\ (86.3)$ & $7.22\ (85.6)$ & $7.76\ (84.5)$ \\
\cite{semihandobj} & Monocular & $6.33\ (87.4)$ & $6.70\ (86.6)$ & $7.17\ (85.7)$ & $8.96\ (82.1)$ \\
\cite{handocc} & Monocular & $\mathbf{5.80}\ (\mathbf{88.4})$ & $\mathbf{6.22}\ (\mathbf{87.6})$ & $\mathbf{6.43}\ (\mathbf{87.2})$ & $\mathbf{7.37}\ (\mathbf{85.3})$ \\
\hline
$S^2HAND(V)$~\cite{tu2023consistent} & Sequence & $7.27\ (85.5)$ & $7.74\ (84.5)$ & $7.71\ (84.6)$ & $7.87\ (84.3)$ \\
VIBE~\cite{vibe} & Sequence & $6.43\ (87.1)$ & $\ 6.72\ (86.5)$ & $6.84\ (86.4)$ & $7.06\ (85.8)$ \\
TCMR~\cite{tcmr} & Sequence & $6.28\ (87.5)$ & $6.56\ (86.9)$ & $6.58\ (86.8)$ & $6.95\ (86.1)$ \\
Ours & Sequence & $\mathbf{5.22}\ (\mathbf{89.6})$ & $\mathbf{5.71}\ (\mathbf{88.6})$ & $\mathbf{5.70}\ (\mathbf{88.6})$ & $\mathbf{6.34}\ (\mathbf{87.3})$ \\ 
\Xhline{1.0pt}
\end{tabular}
\caption{Quantitative MPJPE in mm and (AUC Score) comparison of state-of-the-art hand pose estimation methods on the DexYCB dataset. Our model \textit{(last row)} greatly reduces the error in all hand-object occlusion levels, especially when the hand is heavily occluded \textit{(last column)}.}
\label{tab:res_dexycb}}
\end{table*}

\subsection{Implementation Details}
\textbf{Motion Discrimination }
Our method is capable of processing an image sequence of arbitrary length. For the purposes of this paper, we empirically selected a sequence length of $T=7$ with a gap of 10 frames between consecutive sampled frames to capture meaningful hand motion over approximately 2-3 seconds, while balancing memory cost and performance considerations.

\textbf{Model}
We use the ResNet-50\cite{resnet} followed by ROIAlign\cite{he2017mask} to extract the initial hand feature with a resolution of $32\times 32$ and a feature size of $256$. For the spatial transformer encoder and decoder, we use $3$ multi-head self-attention layers. Each self-attention has $8$ heads and a feedforward dimension of $256$. We use a size of $256$ for the query vector in the transformer decoder. The temporal transformer structure is identical to the spatial transformer.

\textbf{Optimization}
Except that the ResNet-50 is pre-trained on ImageNet, both the SpatioTemporal transformer and motion discriminator are trained from scratch in an end-to-end manner. We use the Adam optimizer with a learning rate of $1\times 10^{-5}$ for the SpatioTemporal transformer, and $1\times 10^{-3}$ for the motion discriminator. Following the standard adversarial training protocol, the generator (SpatioTemporal transformer) and the discriminator are updated alternatively in each step. The entire training takes $60$ epochs and all the learning rate is scaled by a factor of $0.7$ after every $10$ epoch.

\textbf{Computational Cost}
We used 8 RTX 2080Ti GPUs to train the model. On a single RTX 2080Ti, the inference speed of the Deformer inferences is 105 FPS or 15 FPS, depending on the stride selected.

%% file: experiments.tex
In this section, we first describe two public datasets used: DexYCB~\cite{dexycb} and HO3D~\cite{ho3d}. Next, we show our method achieves state-of-the-art performance compared to competitive baselines. Finally, we demonstrate the effectiveness of major designs in ablation studies.

\subsection{Dataset}

\textbf{DexYCB } contains multi-camera videos of
humans grasping 20 YCB~\cite{ycb} objects. It consists of 1000 sequences with 582K frames captured by cameras from 8 different views. During the data collection, there are multiple randomly chosen objects placed on the table. Together with the hand-held object, the hands are frequently occluded during hand-object interaction. We deliberately use these challenging scenarios to test the robustness of the proposed method, especially under occlusions. 

\textbf{HO3D } is a large-scale video dataset for hand pose estimation. It consists of 68 sequences captured from 10 subjects manipulating 10 kinds of YCB~\cite{ycb} objects. There are 77,558 frames: 66,034 for training and 11,524 for testing. As the test set annotation is not released, we submit our predictions to the official online system for evaluation.

\subsection{Evaluation Metric}

We employ the standard metric for hand pose estimation: the Mean Per Joint Position Error (MPJPE) in millimeters (mm), measured after Procrustes alignment. In addition, we present the F-scores and AUC scores as provided by the official evaluation system for each dataset. The released DexYCB test set is further partitioned into three categories based on the hand-object occlusion level, which is determined by the percentage of the hand obscured by the object when re-projected onto the 2D image plane. These levels are set at 25\%, 50\%, and 75\%. We report the results for each category separately to assess the robustness of various methods in handling occlusions. 

Following VIBE~\cite{vibe}, we add a temporal consistency metric: the acceleration error in $(mm)/s^2$. This metric quantifies the discrepancy in acceleration between the ground truth and the predicted 3D joint positions. Moreover, we include the root-aligned MPJPE metric, which evaluates accuracy by aligning the predicted and ground-truth 3D hand roots solely through translation.

\begin{figure}[t]
    \centering
    \includegraphics[width=0.9\linewidth]{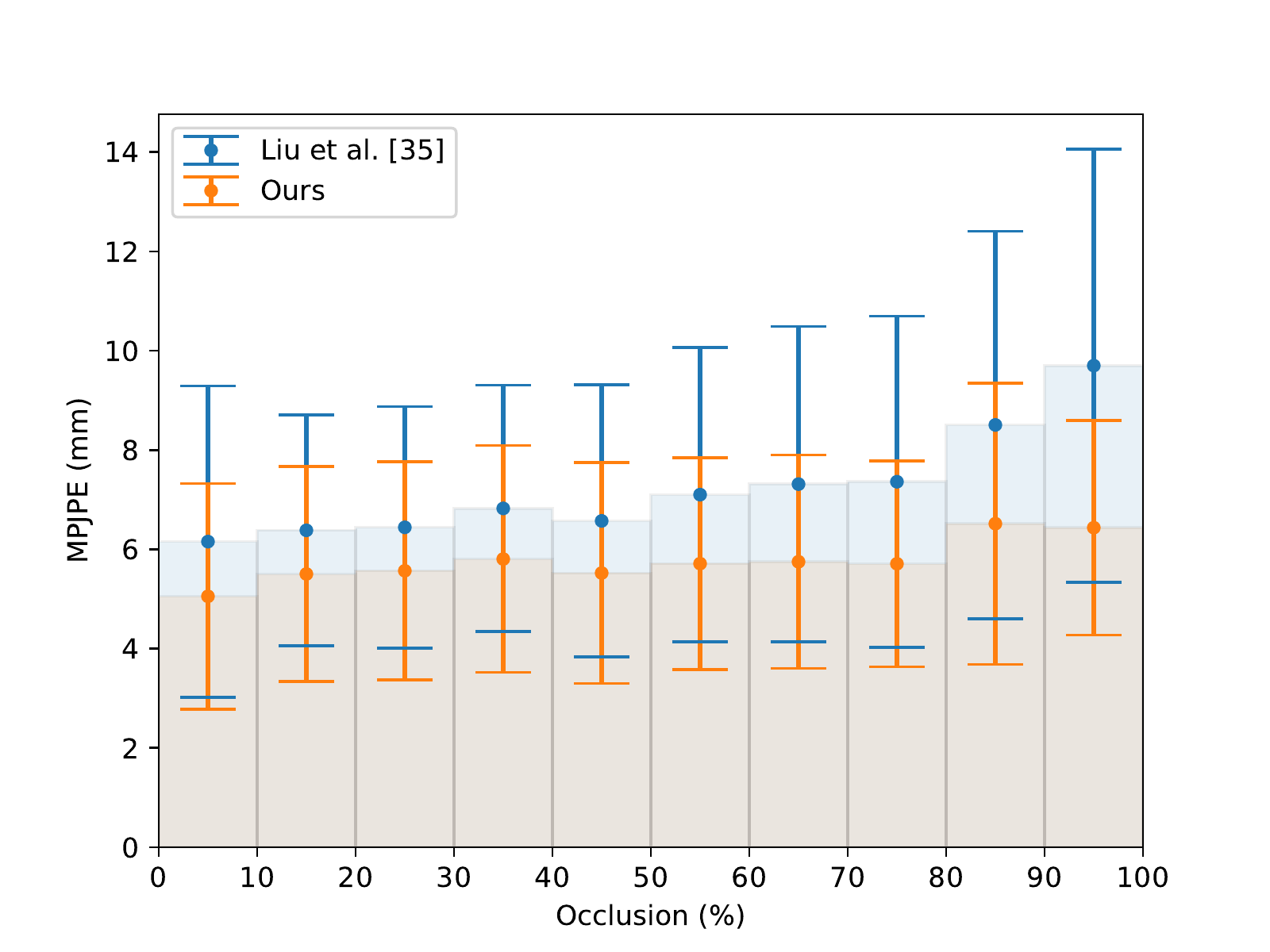}
    \caption{The mean $\pm$ standard deviation of MPJPE on DexYCB test data samples within different hand-object occlusion level ranges. Compared to~\cite{semihandobj}, our method significantly reduces the hand pose estimation error in all occlusion levels, especially when the hand is heavily occluded.}
    \label{fig:coverage_vs_mpjpe}
\end{figure}

\begin{table}[t]
\centering
\resizebox{\linewidth}{!}{
\begin{tabular}{c|cc}
\Xhline{1.0pt}  
 Methods & root-aligned MPJPE & $Accel\ (mm/s^2)$\\
\Xhline{1.0pt}
VIBE~\cite{vibe} & $16.95\ (67.5)$ & $36.4$\\
TCMR~\cite{tcmr} & $16.03\ (70.1)$ & $34.3$\\
$S^2HAND(V)$~\cite{tu2023consistent}  & $19.67\ (62.5)$ & $41.6$\\
MeshGraphormer~\cite{meshformer} & $16.21\ (69.1)$ & $35.9$\\
\cite{semihandobj}+Temporal Transformer+maxMSE & $14.81\ (72.0)$ & $33.3$ \\
\cite{handocc}+Temporal Transformer+maxMSE & $15.15\ (70.8)$ & $33.4$ \\
Ours  & $\mathbf{13.64}$\ ($\mathbf{74.0}$) & $\mathbf{31.7}$\\
\Xhline{1.0pt}
\end{tabular}}
\caption{Quantitative comparison of root-aligned MPJPE in mm and (AUC Score), and the acceleration error on the DexYCB.}
\label{tab:add_metric}
\end{table}

\begin{table}[t]
\centering
\small{
\begin{tabular}{cc|cccc}
\Xhline{1.0pt}  
\multirow{2}{*}{Methods} & \multirow{2}{*}{Input} & \multicolumn{2}{c}{Hand Error ($\downarrow$)} & \multicolumn{2}{c}{Hand F-score ($\uparrow$)}\\
 & & Joint & Mesh & F@5 & F@15\\
\Xhline{1.0pt}
\cite{jointho} & Monocular & $11.1$ & $11.0$ & $46.0$ & $93.0$ \\
\cite{ho3d} & Monocular & $10.7$ & $10.6$ & $50.6$ & $94.2$ \\
\cite{semihandobj} & Monocular & $10.1$ & $9.7$ & $53.2$ & $95.2$ \\
\cite{handocc} & Monocular & $\mathbf{9.1}$ & $\mathbf{8.8}$ & $\mathbf{56.4}$ & $\mathbf{96.3 }$ \\
\hline
VIBE~\cite{vibe} & Sequence & $9.9$ & $9.5$ & $52.6$ & $95.5$ \\
TCMR~\cite{tcmr} & Sequence & $11.4$ & $10.9$ & $46.3$ & $93.3$ \\
{\footnotesize TempCLR~\cite{tempclr}} & Sequence & $10.6$ & $10.6$ & $48.1$ & $93.7$ \\
Ours & Sequence & $\mathbf{9.4}$ & $\mathbf{9.1}$ & $\mathbf{54.6}$ & $\mathbf{96.3}$ \\
\Xhline{1.0pt}
\end{tabular}
\caption{Quantitative MPJPE in mm and F-scores comparison of state-of-the-art hand pose estimation methods on the HO3D dataset. Note that our approach has $31$M parameters, which is 20\% less than \cite{handocc} ($39$M), but still achieves similar performance in the HO3D dataset and better performance in the larger DexYCB dataset.}
\label{tab:res_ho3d}}
\end{table}

\subsection{Results and Analysis}

We compare the proposed method with state-of-the-art video-based method~\cite{tcmr, tu2023consistent, tempclr, vibe}, and single-view methods \cite{a2j, jointho, semihandobj, ho3d, meshformer, handocc, weaklysuphpe} as shown in \cref{tab:res_dexycb},  \cref{tab:add_metric}, and \cref{tab:res_ho3d}. The quantitative results demonstrate our method outperforms previous methods, especially under occlusion cases (also shown in \cref{fig:coverage_vs_mpjpe}). Note that our approach has $31$M parameters, which is 20\% less than \cite{handocc} ($39$M), but still achieves similar performance in the HO3D dataset and better performance in the larger DexYCB dataset.
We also include the error of each hand joint at \cref{fig:mse_vs_maxmse}, where we can observe the proposed maxMSE loss mitigates the error imbalance issue and helps the network focus on critical hand parts like fingertips.

We also provide qualitative comparisons between state-of-the-art methods by visualizing the hand mesh prediction aligned with images in \cref{fig:qualitative_comparsion}. As we shall observe, after addressing both spatial and temporal relationships, our method can robustly produce hand mesh estimations that align with visual evidence and have a more smooth and more plausible hand motion. We include more qualitative results and visualization in the supplementary.

\subsection{Ablation Study}
We perform ablation studies on both HO3D and DexYCB to explain how each component contributes to the final performance. Specifically, we examine the effects of the Dynamic Fusion Module, the maxMSE loss, and the motion discrimination.

\begin{table*}[t]
\centering
\small{
\begin{tabular}{c|cyoe}
\Xhline{1.0pt}  
Aggregation  & All & Occlusion (25\%-50\%) & Occlusion (50\%-75\%) & Occlusion (75\%-100\%)\\
\Xhline{1.0pt}
Center & $5.72\ (88.6)$ & $6.11\ (87.8)$ & $6.14\ (87.7)$ & $6.75\ (86.5)$ \\
Average & $5.40\ (89.2)$ & $5.79\ (88.4)$ & $5.84\ (88.3)$ & $6.39\ (87.2)$ \\
Weighted (Occlusion Level) & $5.34\ (89.3)$ & $5.78\ (88.4)$ & $5.76\ (88.5)$ & $\mathbf{6.33}\ (\mathbf{87.3})$ \\
Dynamic & $\mathbf{5.22}\ (\mathbf{89.6})$ & $\mathbf{5.71}\ (\mathbf{88.6})$ & $\mathbf{5.70}\ (\mathbf{88.6})$ & $6.34\ (\mathbf{87.3})$ \\ 
\Xhline{1.0pt}
\end{tabular}
\caption{Ablation studies of Dynamic Fusion Module on the DexYCB dataset. The evaluation metric is MPJPE in mm and (AUC Score).}
\label{tab:ablation_dynamic_dexycb}}
\end{table*}

\begin{table*}[tb]
\centering
\small{
\begin{tabular}{cc|cyoe}
\Xhline{1.0pt}  
maxMSE & Motion Discrimination  & All & Occlusion (25\%-50\%) & Occlusion (50\%-75\%) & Occlusion (75\%-100\%)\\
\Xhline{1.0pt}
\xmark & \xmark & $5.72\ (88.6)$ & $6.11\ (87.8)$ & $6.14\ (87.7)$ & $6.75\ (86.5)$ \\
\xmark & \cmark & $5.63\ (88.7)$ & $6.00\ (88.0)$ & $5.92\ (88.2)$ & $6.62\ (86.8)$ \\
\cmark & \xmark & $5.43\ (89.1)$ & $6.05\ (87.8)$ & $6.00\ (88.0)$ & $6.59\ (86.8)$ \\
\cmark & \cmark & $\mathbf{5.22}\ (\mathbf{89.6})$ & $\mathbf{5.71}\ (\mathbf{88.6})$ & $\mathbf{5.70}\ (\mathbf{88.6})$ & $\mathbf{6.34}\ (\mathbf{87.3})$ \\ 
\Xhline{1.0pt}
\end{tabular}
\caption{Ablation studies of the maxMSE loss and Motion Discrimination on the DexYCB dataset. The evaluation metric is  MPJPE in mm and (AUC Score).}
\label{tab:ablation_loss_md_dexycb}}
\end{table*}

\begin{table}[t]
\centering
\small{
\begin{tabular}{c|ccccc}
\Xhline{1.0pt}  
\multirow{2}{*}{Aggregation}  & \multicolumn{2}{c}{Hand Error ($\downarrow$)} & \multicolumn{2}{c}{Hand F-score ($\uparrow$)}\\
 & Joint & Mesh & F@5 & F@15\\
\Xhline{1.0pt}
Center & $9.8$ & $9.3$ & $52.8$ & $96.2$ \\
Average & $9.9$ & $9.5$ & $52.6$ & $95.9$ \\
{\footnotesize Weighted (Occlusion Level)} & $9.6$ & $9.2$ & $52.9$ & $96.2$ \\
Dynamic & $\mathbf{9.4}$ & $\mathbf{9.1}$ & $\mathbf{54.6}$ & $\mathbf{96.3}$ \\
\Xhline{1.0pt}
\end{tabular}
\caption{Ablation studies of Dynamic Fusion Module on the HO3D dataset. The evaluation metric is MPJPE in mm and F-scores.}
\label{tab:ablation_dynamic_ho3d}}
\end{table}

\textbf{Effect of Dynamic Fusion Module }
The function of the dynamic fusion module is to explicitly synthesize the hand mesh prediction of neighborhood frames to estimate a more accurate hand mesh at the current frame which is robust to occlusions. It adaptively chooses to focus on specific frames by predicting a weight factor for soft fusion. The dynamic fusion module resolves the ambiguity when the current visual feature is badly contaminated by occlusions or motion blur. To validate its effectiveness, we compare with three variants in \cref{tab:ablation_dynamic_dexycb} and \cref{tab:ablation_dynamic_ho3d}. First, we simply remove this module and evaluate the frame-wise predictions, noted as "Center" in the tables. Secondly, we averaged the deformed predictions from all frames to confirm the benefit of soft fusion, noted as "Average" in the tables. Third, we show the results of using the object occlusion level as the weighting factor, which shows a learnable confidence score that allows the network to take more practical factors other than occlusions into comprehensive consideration.

\begin{figure}[t]
\centering
\includegraphics[width=\hsize]{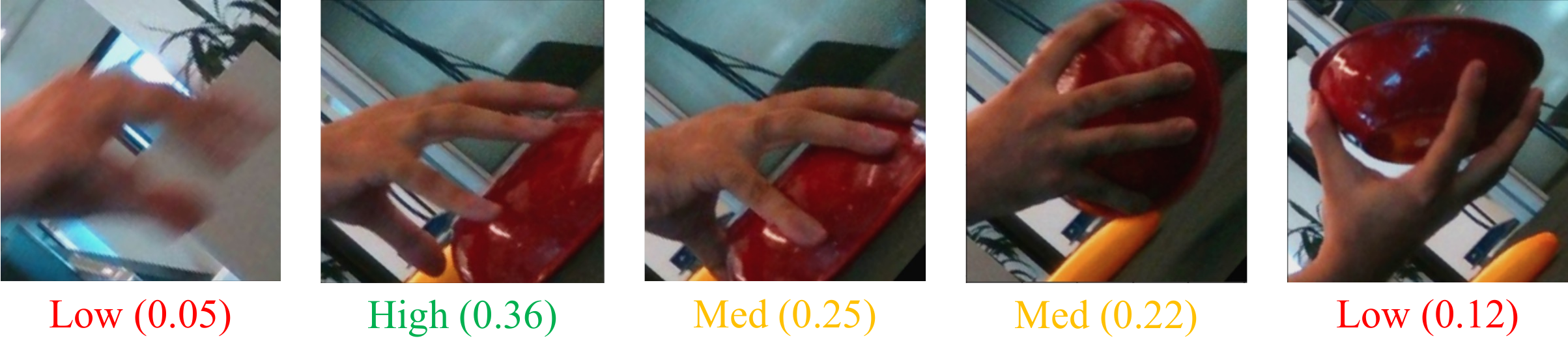}
\caption{Visualization of the confidence score predicted by the Dynamic Fusion Module. The proposed module can implicitly learn visual accountability and assign lower confidence to frames where hands are blurred and occluded.}
\label{fig:confidence}
\end{figure}

\textbf{Effect of the maxMSE Loss }
Compared to vanilla MSE, the proposed maxMSE loss enforces the model to focus on critical hand parts, such as fingertips, by assigning higher weights to the vertices with larger errors. To evaluate the effectiveness of the maxMSE loss, we compare it with a model trained using the standard MSE loss but with an identical architecture.  The quantitative results are presented in \cref{tab:ablation_loss_md_dexycb}, where we can observe the maxMSE leads to better overall performance. Moreover, we visualize the error of every hand joint in \cref{fig:mse_vs_maxmse}, where we observe that the errors around the fingertips are lower and the overall error distribution is more balanced among different hand parts when using the maxMSE loss.

\textbf{Effect of Motion Discrimination } The motion discriminator implemented by a neural network is used to provide adversarial supervision to ensure the predicted hand pose sequence aligns with the data priors. We examine its effectiveness by reporting the performance of the model trained with and without the motion discriminator. \cref{tab:ablation_loss_md_dexycb} reveals the benefits of incorporating adversarial training. 

%% file: conclusion.tex
In this paper, we propose a novel Deformer architecture for robust and plausible 3D hand pose estimation in videos. Our method reasons non-local relationships between hand parts in both spatial and temporal dimensions. To mitigate the issue when a hand is badly occluded or blurred in one frame, we introduce a Dynamic Fusion Module that explicitly and adaptively synthesizes deformed hand mesh prediction of neighborhood frames for robust estimation. Finally, we invented a new loss function, named maxMSE, which adjusts the weight of joints to enforce the model to focus on critical hand parts. We carefully examine every proposed design and demonstrate how they contribute to our state-of-the-art performance on large-scale real-life datasets. Future work shall explore ways to learn fine-grained representation from videos, and reduce the dependency on annotations.

%% file: supplementary.tex
\section*{Supplementary}

\section{Overview}
\label{apdx:overview}
This document provides additional implementation and experimental details, as well as qualitative results and analysis. 
We illustrate the motion discriminator architecture in \cref{apdx:md}. Then we show additional qualitative results in \cref{apdx:qualitative}. Finally, we discuss the limitations of our approach in \cref{apdx:limitations}.

\section{Motion Discriminator}
\label{apdx:md}
\begin{figure}[t]
\centering
\includegraphics[width=0.8\linewidth]{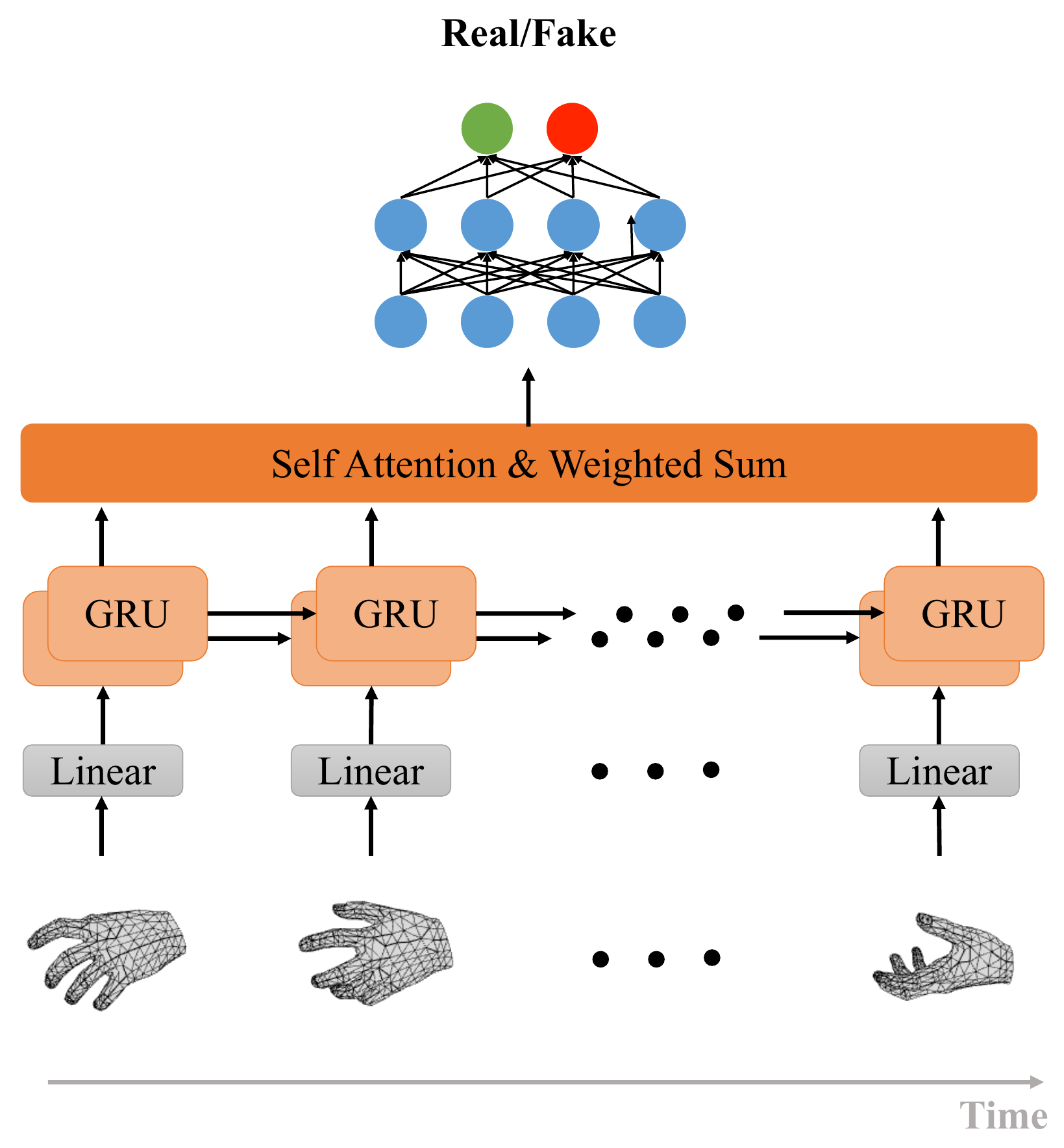}
\caption{The architecture of motion discriminator $\mathcal{D}$. Given a hand pose sequence represented by MANO parameters, $\mathcal{D}$ uses two GRU layers with self-attention to classify if the input is a realistic or fake sequence.}
\label{fig:discriminator}
\end{figure}

As described in  \textit{Sec. 3.2} of the main paper, the motion discriminator $\mathcal{D}$ is learned to supervise the output sequence of the SpatioTemporal Transformer. The architecture of $\mathcal{D}$ is depicted in \cref{fig:discriminator}. Given a hand pose sequence, the motion discriminator first uses a shared linear layer to map the hand pose (represented by MANO parameters) of each timestamp to a high-dimensional feature vector. These frame-wise hand pose features are then input to a two-layer bidirectional Grated Recurrent Unit (GRU), which outputs a new sequence of features that incorporate the information of previous and future hand poses. Instead of using an average or max pooling, we use the self-attention \cite{transformer} mechanism to adaptively choose the most important features in the sequence and summarize them into a single hidden feature. Finally, a linear layer is learned to predict a value in $[0, 1]$ indicating if the input hand pose sequence is realistic or fake.

\section{Qualitative Results}
\label{apdx:qualitative}

\begin{figure}[t]
    \centering
    \includegraphics[width=\linewidth]{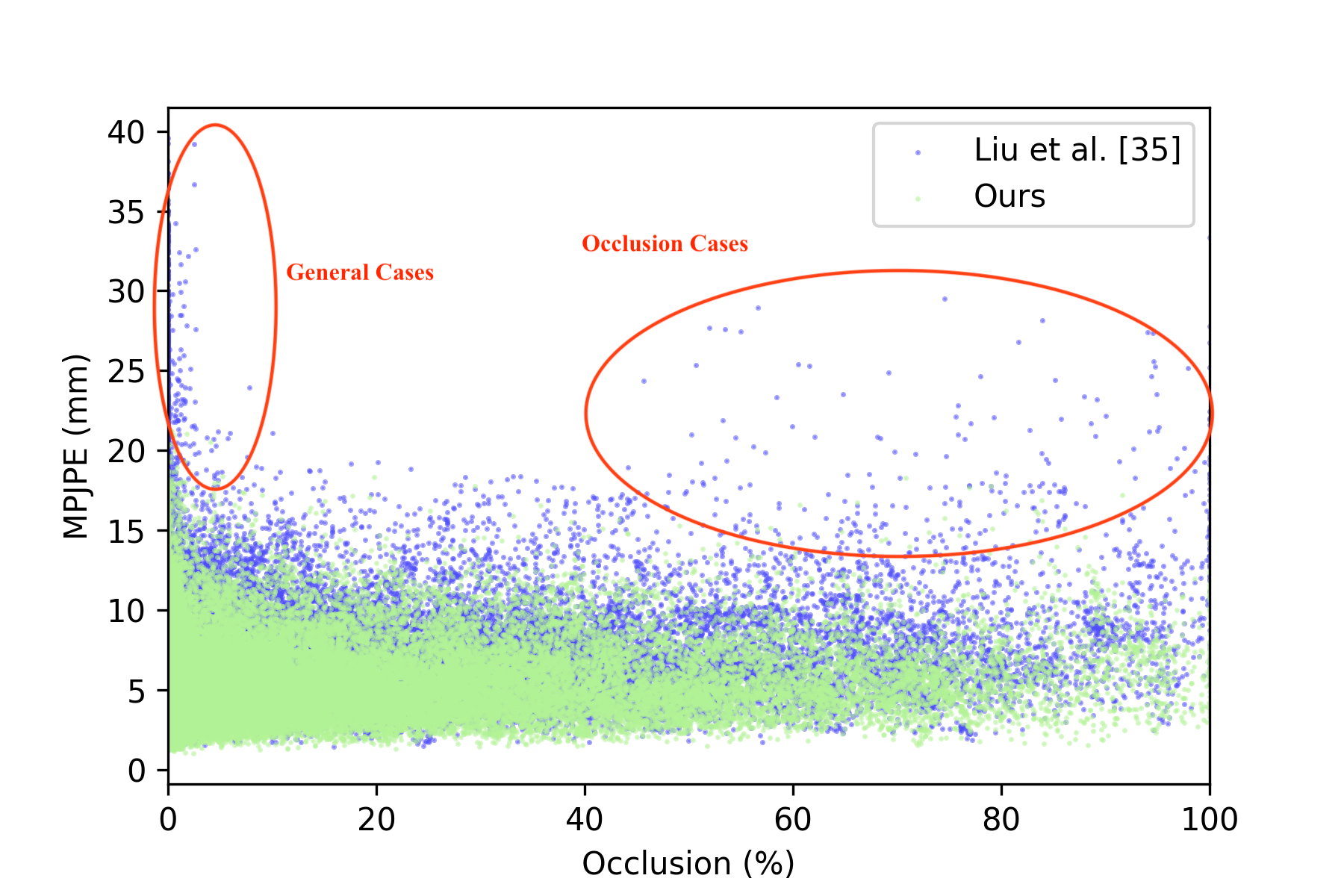}
    \caption{The scatter plot of MPJPE in mm vs. different hand-object occlusion levels on DexYCB test data samples. Compared to~\cite{semihandobj}, our method significantly reduces the hand pose estimation error, especially in occlusion cases.}
    \label{fig:coverage_vs_mpjpe_scatter}
\end{figure}

\begin{figure*}[t]
\centering
\includegraphics[width=0.95\linewidth]{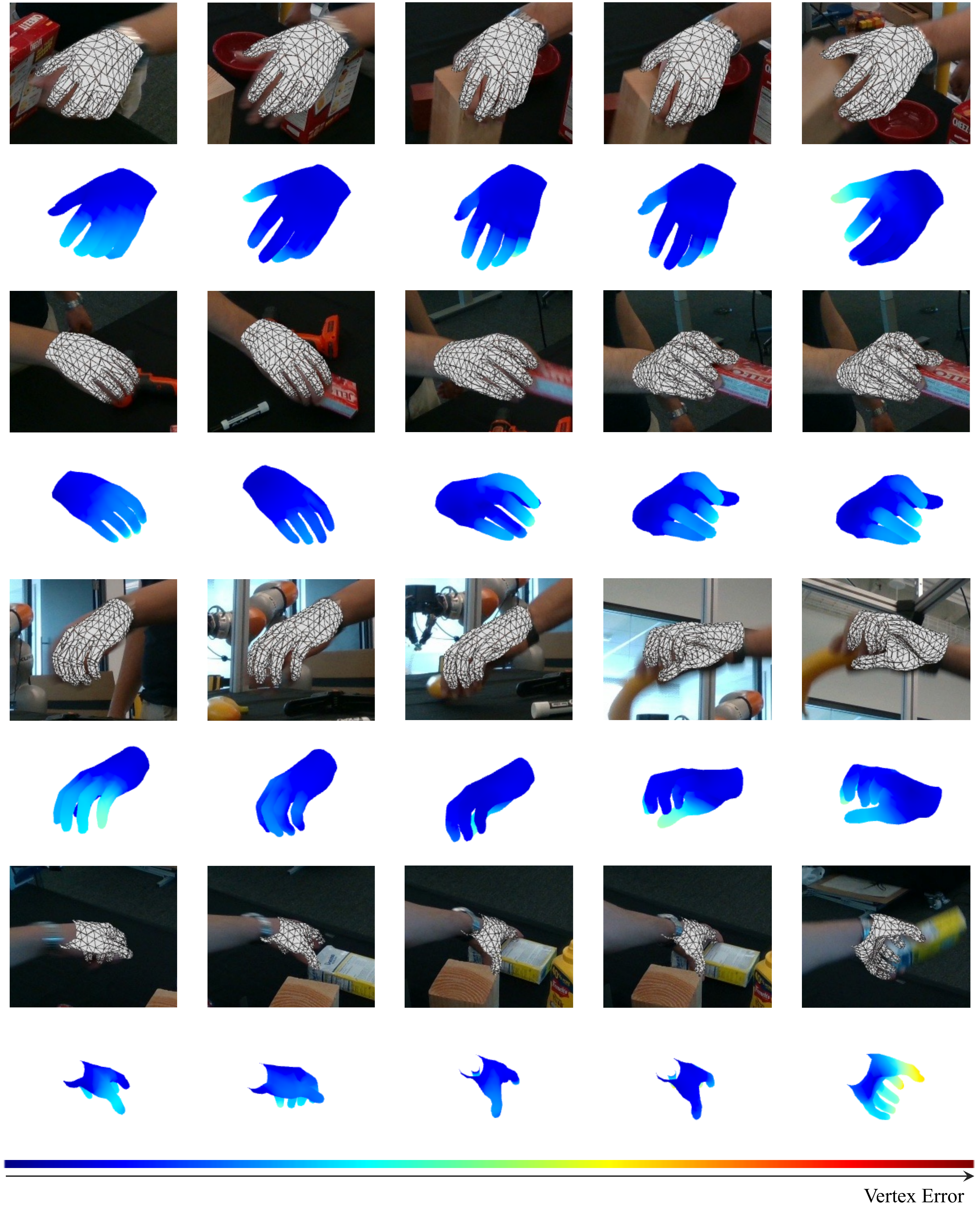}
\caption{Qualitative results on the DexYCB\cite{dexycb} dataset. For every sequence sample, the first row shows the predicted hand mesh and the second row shows the error map.}
\label{fig:qualitative_dexycb}
\end{figure*}

\begin{figure*}[t]
\centering
\includegraphics[width=0.95\linewidth]{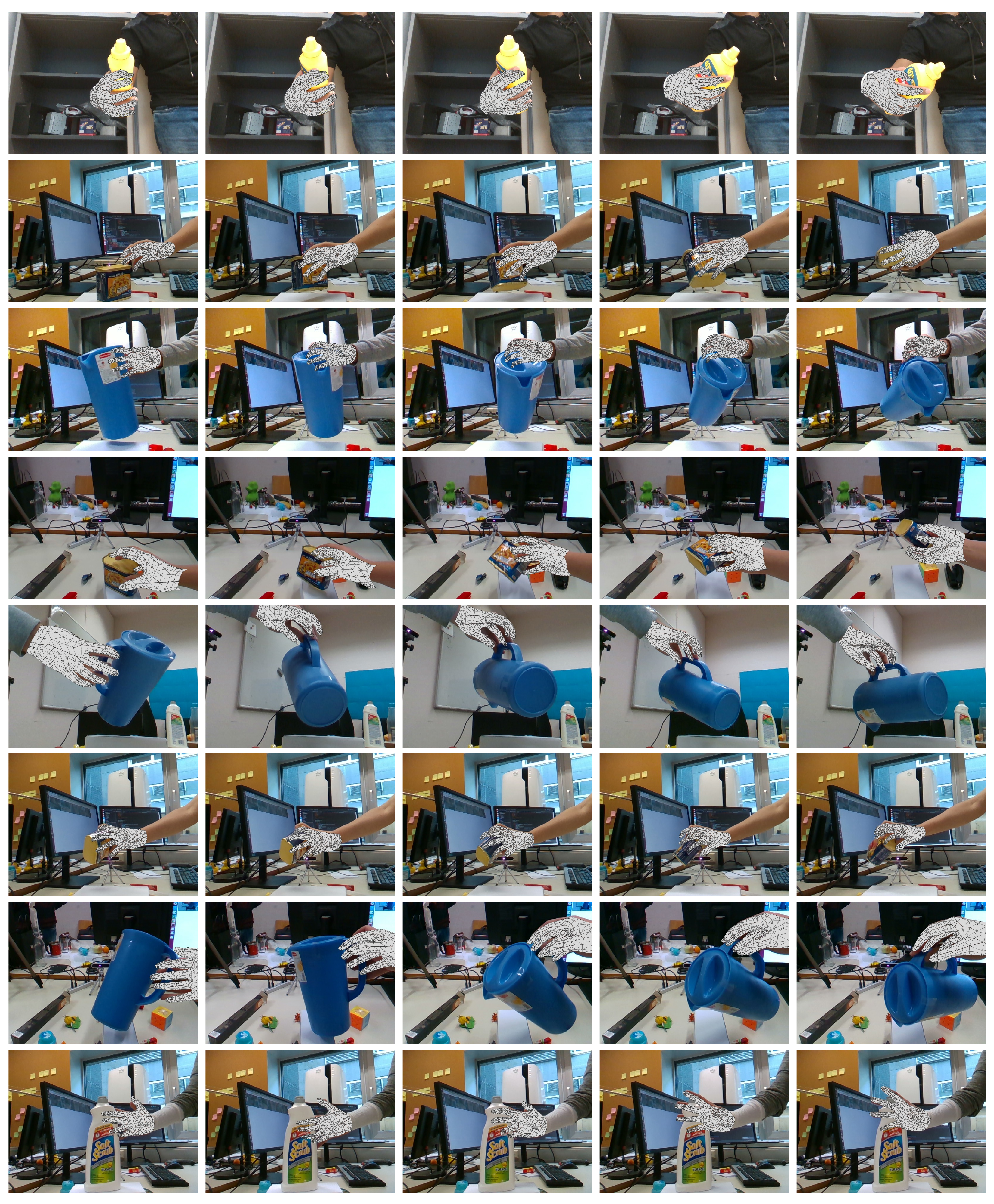}
\caption{Qualitative results on the HO3D\cite{ho3d} dataset. As the HO3D test set annotation is not public, we only show the predicted hand meshes for every exemplary sequence in each row. }
\label{fig:qualitative_ho3d}
\end{figure*}

In this PDF, we present an exemplary set of our hand pose estimation sequence on the DexYCB \cite{dexycb} dataset in \cref{fig:qualitative_dexycb} and on the HO3D\cite{ho3d} dataset in \cref{fig:qualitative_ho3d}. For the DexYCB dataset, each example sequence is visualized in two rows, where the first row shows the predicted hand mesh aligned with the image and the second row shows the error map. For the HO3D dataset, as its test set ground truth is not public, the error map could not be computed and we only show the predicted mesh. The qualitative results and \cref{fig:coverage_vs_mpjpe_scatter} demonstrated that our method is more robust and can estimate accurate hand pose under scenes where the hand is visually occluded or blurred. Most errors are located near the hand parts which are invisible (along all frames in the sequence). In certain cases, another factor that constrains the network is that the mesh generated from MANO parameters (with 778 vertices) is not fine enough to represent the detail of hands. 

\section{Limitations}
\label{apdx:limitations}
First, our method is supervised, which relies on an annotated video dataset to train. Second, the MANO hand model we used has a limited number (778) of vertices. We found in certain cases it fails to capture the details of various hands. Finally, as we used the self-attention mechanism to capture the temporal information, the memory of the proposed method is quadratically proportional to the length of the sequence. This fact limits our approach to efficiently capture the long-term hand motion over the whole video.

